\documentclass{article}

\usepackage[english]{babel}

\usepackage{float}
\usepackage{bbm}
\usepackage{algorithmic}
\usepackage{algorithm}
\usepackage{booktabs} 
\usepackage{multirow}
\usepackage{makecell}
\usepackage[numbers,sort&compress]{natbib}

\usepackage[letterpaper,top=2cm,bottom=2cm,left=3cm,right=3cm,marginparwidth=1.75cm]{geometry}

\usepackage{amsmath}
\usepackage{graphicx}
\usepackage[colorlinks=true, allcolors=blue]{hyperref}

\title{A Clean-graph Backdoor Attack against Graph Convolutional Networks with Poisoned Label Only}
\author{Jiazhu Dai, Haoyu Sun}
\date{}
\begin{document}
\maketitle

\begin{abstract}
Graph Convolutional Networks (GCNs) have shown excellent performance in dealing with various graph structures such as node classification, graph classification and other tasks. However, recent studies have shown that GCNs are vulnerable to a novel threat known as backdoor attacks. However, all existing backdoor attacks in the graph domain require modifying the training samples to accomplish the backdoor injection, which may not be practical in many realistic scenarios where adversaries have no access to modify the training samples and may leads to the backdoor attack being detected easily. In order to explore the backdoor vulnerability of GCNs and create a more practical and stealthy backdoor attack method, this paper proposes a \textbf{c}lean-graph \textbf{b}ackdoor \textbf{a}ttack \textbf{a}gainst GCNs (CBAG) in the node classification task,which only poisons the training labels without any modification to the training samples, revealing that GCNs have this security vulnerability. Specifically, CBAG designs a new trigger exploration method to find important feature dimensions as the trigger patterns to improve the attack performance. By poisoning the training labels, a hidden backdoor is injected into the GCNs model. Experimental results show that our clean graph backdoor can achieve 99\% attack success rate while maintaining the functionality of the GCNs model on benign samples.

\end{abstract}

\section{Introduction}

Graph is ubiquitous in real life, and all kinds of complex relationships in real life can be represented by graphs, for example, social networks\cite{hamilton2017inductive}, molecular structures\cite{irwin2012zinc}, urban transportation networks\cite{xie2019sequential}, biological networks\cite{hirst1992prediction}, etc. In order to efficiently process graph-structured data, various powerful graph neural network models have been proposed. Graph neural networks use the message-passing mechanism to update the representation of nodes by aggregating information from their neighbors. With the rapid development of deep learning, graph convolutional neural networks(GCNs)\cite{kipf2016semi}, as an effective graph neural network model, have achieved great success in graph-structured data processing with powerful graph-structured data learning capability. GCNs make great progress in various tasks on graphs, such as graph classification, node classification, and link prediction.

Despite the great success of GCNs in processing graph-structured data, recent research has shown that GCNs are vulnerable to a security threat known as backdoor attacks\cite{li2022backdoor}. In backdoor attacks, the attacker implants specific patterns (called backdoor triggers) by manipulating a portion of the training data to then embed the backdoor triggers into the GCNs during the training phase. In the model inference phase, when the input sample contains the backdoor trigger, the backdoor in the model will be activated and perform the action specified by the attacker in advance, e.g., the backdoor model incorrectly classifies that input sample with the backdoor trigger to the label specified by the attacker(called the target label), while for the samples that do not contain the backdoor trigger, the model with implanted backdoor behaves normally. Backdoor attacks usually occur when the model training process is uncontrolled, e.g., backdoor attacks occur when using models trained by third parties or using third-party data to train models. Backdoor attacks are highly stealthy and expose GCNs to serious security challenges.

There has been some studies \cite{xi2021graph,zhang2021backdoor,xu2021explainability,dai2023unnoticeable,zheng2023motif,yang2022transferable,chen2023feature,dai2023semantic,yang2023percba}on graph backdoor attacks. However, the existing graph backdoor attacks usually have a prerequisite assumption that the attacker can have the ability to modify the structure or features of the training data or inject additional nodes and edges to inject triggers, for example, zhang et al\cite{zhang2021backdoor} by modifying a part of the structure of the training data to a fixed subgraph structure as the backdoor trigger . xu et al\cite{xu2021explainability} use an interpretable method to find important feature dimensions, and then modify a portion of the nodes' features to fixed values as the backdoor trigger. Dai et al\cite{dai2023unnoticeable} used additional subgraphs as the backdoor trigger by injecting additional subgraphs into the training graph to implement an unnoticeable backdoor attack. However, in practical situations, the attacker may often not have access to modify the training samples, for example, the current training of artificial intelligence models relies on a large amount of manual labeling, and the data of the training model is outsourced to a third-party staff to label the data. The attacker will not be able to modify the original training samples to inject triggers, so the attacker will not be able to use the existing backdoor attack methods to implement backdoor attacks. But, the attacker can disguise himself as the staff member who labeled the training data to modify the labels of the samples. Therefore, in this paper, we propose an interesting question: is it possible to accomplish a backdoor attack on the graph by modifying only the labels of the nodes?

The answer is affirmative. GCNs depend on the structural and feature information of the graph to accomplish node classification, and each node has its own feature information. Some combinations of features can be regarded as the trigger of backdoor attack, for other nodes with information of these combinations of features, we select the appropriate nodes to poison, and implement backdoor attack on the victim model by modifying the labels of these nodes to the target label, which affects the misclassification of the model to the target label.

Therefore, in this paper, we design a novel \textbf{c}lean-graph \textbf{b}ackdoor \textbf{a}ttack against \textbf{G}CNs (CBAG) in the context of node classification, which only poisons the training labels without making any modification to the graph structure and features during the training process. Specifically, we select the robust features combination that are strongly associated with the target label as the trigger pattern. The appropriate nodes are then selected for labeled poisoning via a scoring function. After label poisoning is completed, the training samples becomes the poisoned dataset with poisoned labels. After training the target model using the poisoned training dataset, the backdoor can then be embedded into the GCN models, which establish an association between the trigger and the target label. Therefore, during the whole poisoning process, CBAG only modified the labels of the training samples to implant the backdoor trigger. In the model inference phase, the backdoor is activated by the attacker by modifying the values of the selected the features combination of samples to be the maximum value of the original features in the graph, and the backdoor GCN model gives the malicious classification result specified by the attacker.

To the best of our knowledge, this is the first Clean-graph backdoor attack on the graph node classification task. Our contribution is as follows:

\begin{itemize}
\item We propose a Clean-graph backdoor attack against GCNs (CBAG) and reveal the existence of this security
vulnerability in GCNs. CBAG has the following features: 1. The attacker does not make any modifications to the original graph structure and features or inject additional nodes and edges into the graph during the training process, and the attacker only needs to modify the labels of the training nodes to implement the backdoor attack. 2. The attacker uses the original features present in the graph to activate the backdoor, which helps to enhance the stealthiness of the attack. 3. The CBAG is a black-box attack, and we assume that the attacker has no priori knowledge about the details of the training of the victim’s model and cannot manipulate the training process of the victim’s mode; they only need to access the training data and manipulate the training process by poisoning some labels.,
\item Detailed experiments based on four different real-world datasets are conducted to evaluate the performances of CBAG, and it performs high attack
success rate (up to 97\%) without distinct model
accuracy degradation on clean data, while the poison
rate is lower than 5\%.
\end{itemize}

The rest of the paper is organized as follows. Firstly, we introduce the background knowledge of GCNs  in Section \ref{section:2}; Then, we introduce the adversarial attacks and backdoor attacks against GCNs in Section \ref{section:3}; Next, we describe in detail our proposed backdoor attack in Section \ref{section:4} and evaluate its performance in Section \ref{section:5};Finally, we conclude the paper in Section \ref{section:6}.

\section{Background}
\label{section:2}
\subsection{Graph Convolutional Networks (GCNs)}
Given an undirected and unweighted attribute graph $G=(V,E,X)$, where $V=\{v_1,v_2,\cdots,v_N\}$ is the set of nodes,  and $ N=|V|$ is the total number of nodes. $E$ is the set of edges. $X=R^{N \times d}$ represents the nodes attribute matrix and $d$ is the dimension of the feature.  $A \in R^{N\times N}$ is the adjacency matrix of the graph. For two nodes $v_i,v_j \in V$, if $(v_i,v_j)\in E$, it means that there exists an edge between $v_i$ with $v_j$ and $A_{ij}=1$, otherwise, $A_{ij}=0$. 

Graph Neural Networks achieve excellent performance in processing graph data, while Graph Convolutional Networks(GCNs), as an effective graph neural network, achieve outstanding performance in various graph-based tasks. Therefore, in this paper, we highly focus on GCNs. Specifically, the GCN updates the representation of each layer through the following propagation rules:
\begin{equation}
H^{k+1}=\sigma(\tilde{D}^\frac{1}{2}\tilde{A}\tilde{D}^{-\frac{1}{2}}H^{k}{W}^k)
\end{equation}
where $H^{k+1}$ denotes the node representation of the k-th layer and the initial node representation $H^{0}=X$. $\sigma(\cdot)$is an activation function such as the ReLU function. $\tilde{A} = A+I$ is a self-connected adjacency matrix of the graph $G$, which allows the GCN to incorporate the node features themselves when updating the node representations. $I\in R^{N \times N}$ is the the identity matrix. $\tilde{D}$ is the diagonal matrix of $\tilde{A}$, where $\tilde{D}_{ii} = \sum_{j}\tilde{A}_{ij}$. $W^k\in R^{F \times F^{'}}$is the k-th layer weight matrix, which will be trained during the optimization. $F$ and $F^{'}$are the node representation dimensions at the k-th and k+1-th layers respectively.


\section{Related work}

In this section, we introduce some work on adversarial attacks and backdoor attacks against GCNs.

\subsection{Adversarial Attacks against GCNs}

Despite the remarkable performance that GCNs have demonstrated across various graph learning tasks and their widespread application, recent research indicates that GCNs are similarly vulnerable to adversarial attacks. According to the stage at which the attack occurs, adversarial attacks can be classified into evasion attacks\cite{chen2020survey,dai2022targeted,dai2018adversarial} and poisoning attacks\cite{sun2020adversarial,zugner2018adversarial,zugner2020adversarial,zhang2020adversarial}. Evasion attacks occur during the testing phase, where GCN models have already been well-trained. In these attacks, adversaries cannot modify the model directly; instead, they add perturbations to target test samples. Poisoning attacks typically occur during the training phase, where attackers manipulate the training graph before training the GCN model. This manipulation often results in the GCN model trained on poisoned datasets exhibiting lower prediction accuracy on test samples.

\subsection{Backdoor Attacks against GCNs}

Recent studies indicate that GCNs, like other neural networks, are vulnerable to backdoor attacks. Backdoor attacks, as a particular type of poisoning attack, were initially introduced in the context of image domains. Gu et al \cite{gu2017badnets} proposed the first image backdoor attack model: BadNets, by using special markers as triggers pasted to the training samples and then the backdoor model is trained on this poisoned training set. In the inference phase, the backdoor model classifies images with triggers as target labels and classifies clean images normally. Subsequently, backdoor attacks in the textual domain\cite{dai2019backdoor} have also been extensively studied. 

In the graph domain, Xi et al\cite{xi2021graph} proposed for the first time a backdoor attack method against GNNs, which uses subgraphs as triggers that can be dynamically customized for different graphs with different triggers to poison the data. Yang et al\cite{yang2022transferable} proposed a transferable graph backdoor attack without a fixed trigger pattern to implement a black-box attack on GNNs through a surrogate model. Chen et al\cite{chen2023feature} utilized the feature subsets of nodes as triggers and modified the node labels to the target classes after implanting the triggers. Since the insertion of the trigger destroys the similarity between the nodes,to ensure that the modification of the labels does not destroy the similarity between the nodes,they adaptively adjusted the structure of the graph. Yang et al\cite{yang2023percba} proposed a Persistent Clean-label Backdoor Attacks on Node Classification, which antagonistically perturbs the features of unmarked nodes to enforce the model to classify them into the premeditated class. Dai et al\cite{dai2023unnoticeable} study a novel problem of unnoticeable graph backdoor attacks, which deliberately select the nodes to inject triggers and target class labels in the poisoning phase adn deploy adaptive trigger generator to obtain effective triggers that are difficult to be noticed.

These already proposed backdoor attack methods above require changes to the training samples or injection of additional nodes and edges, and this behavior which requires modification of the training samples can be easily detected by the victim, whereas CBAG considers a more realistic and stealthy attack scenario, i.e., only modifying the label to complete the injection of the backdoor. Similar to CBAG, adversarial label-flipping attack\cite{zhang2020adversarial} also accomplishes the poisoning attack by poisoning only the labels without changing the training samples, which has been accomplished effectively on GNN. The advantage of these attacks is that they do not modify the training samples, making this attack more stealthy. Similarly, CBAG operates only on the labels of the training samples, which makes our attack more stealthy than existing backdoor attacks that rely on perturbation graphs. But unlike adversarial label-flipping attack, which reduces the accuracy of the model by flipping the labels, CBAG is a backdoor attack that aims to manipulate the victim model in a controlled way by adding triggers to the victim's model.

\label{section:3}
\begin{figure}[htbp]
\centering
\includegraphics[width=1\linewidth]{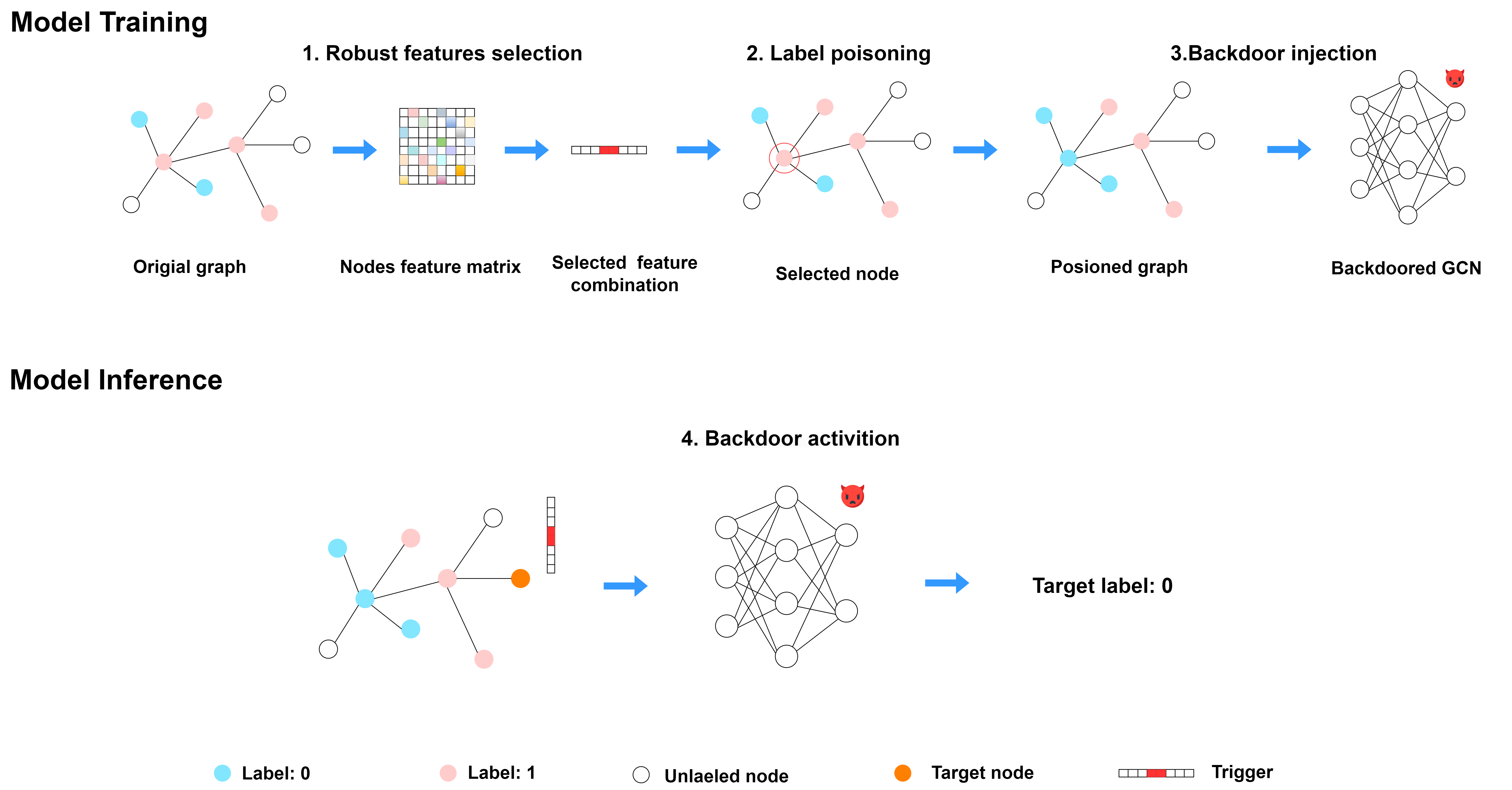}
\caption{\label{fig:frame}The framework of CBAG.}
\end{figure}

\section{A Clean-graph Backdoor Attack against GCNs}
\label{section:4}

In this section, we illustrate in detail how CBAG is implemented.Table \ref{tab:3} summarizes the the notions used in the following sections and their explanations.

\begin{table}[htbp]
\centering
\caption{\label{tab:3}The explanations of the notions}
\begin{tabular}{cc}
\toprule
Notation & Explanations \\
\midrule
$G=(V,E,X)$&Graph $G$ with edge set $E$, node set $V$ and feature matrix $X$ \\
$\lvert V \rvert$&The number of nodes \\
$f_c$&The clean GCN model\\
$f_b$&The backdoored GCN model\\
$f_s$&A scoring function for rating the candidate samples\\
$y_t$&\makecell[c]{Target label, which is the label specified by the attacker, who intends\\ for all samples with the trigger to be predicted by the backdoor model}\\
$k$&The number of the features combination\\
$m$&The number of target label nodes in the training set\\
$N$&The number of nodes in the training set\\
$\mathbbm{I}(\cdot)$& The indicator function.\\

\bottomrule
\end{tabular}
\end{table}

\subsection{Attack Overview}

In traditional node classification backdoor attacks, the attacker selects poisoned nodes in the training samples, and then tends to modify each node to inject the backdoor in a fixed way, such as injecting the same feature value in the feature subset of each poisoned node\cite{xu2021explainability,chen2023feature} or injecting the subgraph consisting of the same number of malicious nodes\cite{dai2023unnoticeable} connecting each poisoned node. This not only requires the attacker to gain access to modify the training set, but also this kind of malicious behavior can be easily detected. Unlike these traditional backdoor attacks, CBAG consider a more practical and realistic scenario: the attacker is unable to make any modifications to the training samples, e.g., changing the structure and features of the original graph to embed triggers or injecting additional nodes and edges to embed triggers during training phase. The attacker can only mislabel the labels of a small fraction of the nodes in the training samples.  This malicious behavior is highly stealthy because it is difficult to distinguish from common labeling errors. For example,MIT study finds "systematic" labeling errors are prevalent in AI benchmark datasets\cite{northcutt2021pervasive},  so it is difficult to detect a backdoor attack that injects triggers by mislabeling a small subset of samples.

We assume that the attacker has no priori knowledge about the details of the training of the victim's model (e.g., model structure and loss function) and cannot manipulate the training process of the victim's model. The attacker can access the training samples and modify the labels of some samples. The objective of the attacker is to train the backdoor model $f_b$, which will misclassify the node $v_t$ with trigger as the target label designed by the attacker, whereas for the normal node $v$, $f_b$ behaves normally. The attacker's objective can be formalized as follows:
\begin{equation}\label{eq1}
\left\{
\begin{aligned}
f_b(v_t)&=y_t,\\
f_b(v)&=f_c(v).\\
\end{aligned}
\right.
\end{equation}
From Equation \ref{eq1}, the first objective shows the effectiveness of the attack, i.e., the backdoor GCN can successfully predict the sample with the trigger as the target label $y_t$, and the second objective shows the evasiveness of the attack, i.e., the backdoor GCN's behavior on clean samples is consistent with that of the clean GCN.

The realization process of CBAG is shown in Fig. \ref{fig:frame} and includes the following steps:
\begin{enumerate}
\item \textbf{Selecting the robust feature combination as the trigger pattern: }The attacker analyzes the features of the nodes in the training set and selects the robust combination of features as the trigger pattern. This will be described in detail in 4.2.
\item \textbf{Label poisoning: }Initially, the attacker selects nodes with non-target label from the training set. Then, the attacker selects the nodes with top-n scores of non-target label based on the score function and relabels them as the target label to generate the poisoned dataset. This will be described in detail in Section 4.3.   
\item \textbf{Backdoor injection: }The target model is trained with the poisoned training samples to embed the backdoor into the GCN model. This will be described in detail in Section 4.4
\item \textbf{Backdoor activation: }After embedding the backdoor in the GCN model, the attacker can activate the backdoor in the model by modifying the value of the combination of features we choose to be the largest value of the original feature in the graph and predict it as the target label. While normal nodes will be predicted normally. This will be described in detail in Section 4.5.                                 
\end{enumerate}
\subsection{Selecting the feature combination as the trigger pattern}
Compared to traditional backdoor attacks on node classification, CBAG does not modify the features and structure of the graph or inject malicious nodes and edges during the training process,and CBAG only need to change training labels in the annotations. In order to implement CBAG, which only needs to poison the labels of the nodes, we need to select a features combination as the trigger pattern, which have a strong association with the target label. In general, the larger the value of a node's features in a graph, the more robust it is, and a node's own robustness features are usually strongly associated with its own Label, but a single node's own feature value is difficult to represent the entire graph. In order to find the robust feature combination that have a strong association with the target label, we firstly calculate the average feature values of all the target label nodes in the training samples. However, some features may not only have high values in the target label nodes, it also has high values in the non-target label nodes, and these features may not have a strong association with the target label, so these features may not be helpful for the target model to establish the association between the trigger and the target label. In order to eliminate the effect of these features, we calculate the average features values of all nodes in the training samples again, and finally we subtract the average features value of all nodes from the average features value of the nodes with the target label, and the difference size reflects the robustness of the features, and the larger the value is, the stronger the robustness is, and the stronger the association with the target label is. Therefore we define this difference as the importance of each feature, and finally we use the combination of features with the largest importance as the trigger pattern, which facilitates the model to establish the association between the trigger and the target label. 

Specifically, we formalize the process of selecting the combination of features. For $\forall v_i \in V$
with a d-dimensional feature vector $x_i=\{x_{i1},x_{i2},\dots,x_{id}\}$. Our features' importance is calculated as follows:
\begin{equation}
    x_s = \frac{1}{m}\sum_{i=1}^{N}x_i\mathbbm{I}(y_i = y_t)-\frac{1}{N}\sum_{i=1}^{N}x_i
\end{equation}

where $m$ is the number of target label nodes, $N$ is the number of nodes in the training set, and $\mathbbm{I}(\cdot)$ is the indicator function. After obtaining the importance of each feature, we sort the values in order of magnitude and select the index combination of the largest feature value as the trigger pattern according to the predefined trigger size.
\subsection{Label poisoning}
Once our trigger pattern is determined, the next step is to select nodes with non-target label to change their labels in the annotations. Among the non-target label nodes, there are some nodes whose features will also have high values in these feature dimensions we have chosen, and the labels of these nodes are not the target labels, which will prevent the model from establishing an association between the trigger and the target label. So in order to select the poisoned nodes, we propose a scoring function $f_s$ to score each non-target label node and select the $n$ non-target label nodes with higher scores as poisoned nodes. We will modify their labels to the target label to complete the poisoning. Specifically, assume that we select a combination of feature indexes as $S=\{s_1,s_2,\dots,s_k\}$, and we sum up the values of each non-target label node in each of these feature dimensions as their scores. The scoring function can be formalized as:
\begin{equation}
    f_s(v_i)=\sum_{s\in S}x_i[s]
\end{equation}

where $x_i [s]$ denotes the feature value of node $v_i$ with inedex $s$. Finally, the $n$ non-target label nodes with the highest scores are selected as poisoned samples and their labels are relabeled as the target labels.

\subsection{Backdoor injection}
Once the label poisoning is completed, the training dataset becomes a poisoned dataset with poisoned samples. After training the GCN model using the poisoned training samples, the backdoor can then be embedded into the GCN model, which establishes the association between the trigger and the target label.

\subsection{Backdoor activation}

Once the infected model is deployed for inference, the attacker can attack it using predefined trigger. Specifically, the attacker will misclassify a test node as the target label by modifying the combination of features we choose to be the maximum of the original features in the graph, and then the infected model will misclassify the test node as the target label. The primitive features are chosen because the primitive features are naturally present in the node's feature vector, which helps to enhance the stealthiness of the attack.

\section{Experiments}
\label{section:5}
In this section, we evaluate the effectiveness of CBAG on the node classification task through three experiments. Firstly, we evaluate the attack success rate of CBAG and the model accuracy difference of the backdoor GCN model compared to the clean GCN model; then we test the effect of different poisoning rates on the performance of CBAG; finally, we test the effect of the size of the number of feature combinations on the performance of CBAG.
\subsection{Experiments setting}
\subsubsection{Datasets}
We employ four frequently used real-world datasets (dataset A: Cora, dataset B: Citeseer, dataset C: Cora\_ML, dataset D: Pubmed) to measure CBAG performance, and
dataset statistics are shown in Table \ref{tab:1}. For each dataset we randomly split the training and validation sets by 20\% each and the remaining nodes are used as the test set.

\begin{table}[htbp]
\caption{\label{tab:1}The dataset statistics}
\centering
\begin{tabular}{ccccccc}
\toprule
Datasets&Nodes&Edges&Classes&Features&Graphs [Class]&target label\\
\midrule
Cora&2708&5429&7&1433&\makecell[c]{351[0], 217[1], 418[2], 818[3],\\
426[4], 298[5], 180[6]}&6\\
CiteSeer&3327&4732&6&3703&\makecell[c]{264[0], 590[1], 668[2],\\ 
701[3], 596[4], 508[5]}&0\\
Cora\_ML&2810&7981&7&2879&\makecell[c]{354[0], 402[1], 452[2], 442[3],\\857[4], 193[5], 295[6]}&5\\
PubMed&19717&44338&3&500&4103[0], 7739[1], 7875[2]&0\\
\bottomrule
\end{tabular}
\end{table}

\subsubsection{Metrics}
In this paper, we mainly use the following evaluation metrics: attack success rate, clean accuracy degradation, and poisoning rate to evaluate the effectiveness of CBAG.
\begin{enumerate}
\item Attack Success Rate (ASR) is the probability that the samples with the trigger are predicted by the backdoor model to be the target label.
\begin{equation}
    ASR = \frac{\sum_{i=1}^{n}\mathbbm{I}(f_b(v_i)=y_t)}{n}
\end{equation}
where $\mathbbm{I}$ represents the indicator function.
\item Clean accuracy drop (CAD) is the difference between the classification accuracy of the clean GCN on benign samples and the classification accuracy of the backdoor GCN on benign samples.
\begin{equation}
    CAD=ACC_c-ACC_b
\end{equation}
where $ACC_c$ and $ACC_b$ are the classification accuracy of the clean and backdoor model, respectively. Lower CAD represents better performance, indicating that the backdoor and clean model are closer in accuracy on clean samples and that the backdoor attack is better stealthy.
\item Poisoning rate ($p$) is the ratio of the number of label poisoning samples to the total number of samples in the training set. The lower the poisoning rate, the more stealthy the backdoor attack is.

\end{enumerate}
\subsubsection{Parameter settings}

We use the GCN model as the target model to validate the effectiveness of the CBAG. The GCN model contains a two-layer GCN structure, where the units of the hidden layer are 32, optimizer is set to Adam, weight decay is $5e^{-4}$, the dropout rate is 0.6, epochs set as 200 and the learning rate is 0.01. 

\subsubsection{Baseline}
Although there are some backdoor attack methods on the node classification task, these methods require modification of the original graph to inject the trigger to complete the backdoor attack. So these methods are not able to perform the backdoor attack if it is in the case of only modifying the label without making any modification to the original graph. Therefore, in order to evaluate the effectiveness of CBAG, we add CBAG-R, a variant of CBAG, which randomly selects a combination of features to be used as the trigger pattern.

\subsection{Experiments results}
\subsubsection{Results on Different Datasets}

To evaluate the effectiveness of CBAG, we evaluate the attacks on the target model on four benchmark datasets.For each attack, the poisoning rate is set to 0.05 and the feature combination size is set to 0.05. The experimental results are shown in Table \ref{tab:2}. The experimental results showed that CBAG achieved excellent performance. For the target model, CBAG achieved 99.83\%, 96.80\%, 91.30\%, and 98.30\% attack success rates on the four datasets, respectively, which indicates that CBAG can successfully accomplish the prediction of the label of the poisoned nodes as the target label. Obviously, the attack success rates of our method are better than CBAG-R on all datasets, which demonstrates that the robust combination of features we choose can better facilitate the target model to establish the association between the trigger and the target label. 

On the other hand, the clean accuracy drop of CBAG on the four datasets is 2.41\%, -1.93\%, 0.15\%, and 1.38\%, respectively, which indicates that the backdoor model's prediction accuracy for clean samples is close to that of the clean model, and further suggests that CBAG has a high attack evasiveness performance. Compared with CBAG-R, on Cora and PubMed datasets, the CAD of CBAG is slightly higher than that of CBAG-R, which may be due to the fact that CBAG achieves a higher attack success rate, resulting in more clean nodes being affected. On the other datasets, the CAD of the two is not much different.

Overall, the CBAG achieves high attack success rates and good stealth at low poisoning rates.


\begin{table}[htbp]
\centering
\caption{\label{tab:2} Comparison with the baseline.}
\begin{tabular}{@{}ccccc@{}}
\toprule
\multirow{2}{*}{Datasets} & \multicolumn{2}{c}{ASR(\%)}   & \multicolumn{2}{c}{CAD(\%)}  \\ \cmidrule(l){2-5} 
                         & \multicolumn{1}{c}{CBAG-R} & \multicolumn{1}{c}{CBAG}  & \multicolumn{1}{c}{CBAG-R} & \multicolumn{1}{c}{CBAG}  \\ \cmidrule(r){1-5}
\multirow{1}{*}{Cora}          &93.60   &99.83     &1.69  &2.41  \\                  
\multirow{1}{*}{CiteSeer}      &53.32   &96.80     &-1.74 &-1.93    \\                   
\multirow{1}{*}{Cora\_ML}      &28.75   &91.30     &0.08  &0.15   \\                 
\multirow{1}{*}{PubMed}        &95.02   &98.30     &0.31  &1.38    \\
\cmidrule(l){1-5}
\end{tabular}
\end{table}

\begin{figure}[htbp]
\centering
\includegraphics[width=1\linewidth]{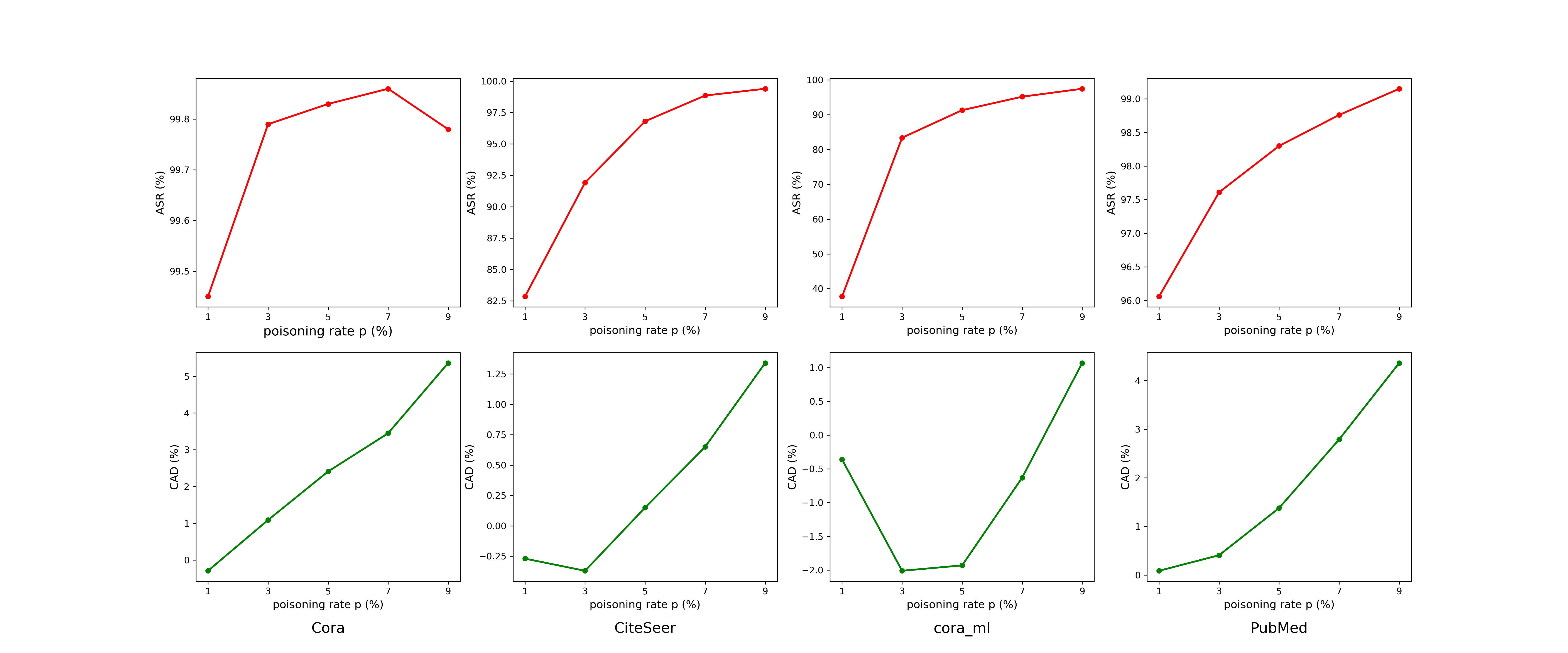}
\caption{\label{fig:rate}The impact of poisoning rates on ASR and CAD.}
\end{figure}

\subsubsection{Impact of poisoning rate}

In this section, we investigate the effect of different poisoning rates on ASR and CAD, as shown in Fig. \ref{fig:rate}. The horizontal axis of each graph represents the poisoning rate, the vertical coordinates of the top four figures represent the ASR, and the vertical coordinates of the bottom four figures represent the CAD. From the figure we can observe that as the poisoning rate increases, the ASR increases accordingly and the same trend is observed in all four datasets. In addition, on the Cora and PubMed datasets, a small poisoning rate can achieve a high ASR, which is enough to launch an effective attack on the target model, while on the Cora\_ML and CiteSeer datasets, the ASRs are 37.76\% and 82.25\% when p=1\%, and the ASRs are increased to 91.30\% and 96.30\% when p=5\%, which indicates that CBAG only needs to poison a small portion of nodes' labels to achieve good attack performance, which can ensure that our attack can be concealed as a common labeling error to implement backdoor attacks. on the other hand, the increase of p has a negative effect on CAD. When p is very small, CAD will also be small, and when p increases, the CAD of Cora and PubMed datasets will also increase as p increases, while the CAD of Cora\_ML and CiteSeer datasets will have a tendency to decrease and then increase, but all of them maintain an overall increasing trend. 

From the above results, we can get the following three conclusions:(1) When the poisoning rate increases to 9\%, the attack success rate of CBAG can be close to 99\%;(2) The prediction accuracy of the backdoor GCN model for benign samples is close to that of the clean model. (3) CBAG can implement backdoor attacks relying on a small attack success rate, which indicates that CBAG can disguise itself as a common labeling error to implement a backdoor attack, which is difficult to be detected by the defender.

\begin{figure}[htbp]
\centering
\includegraphics[width=1\linewidth]{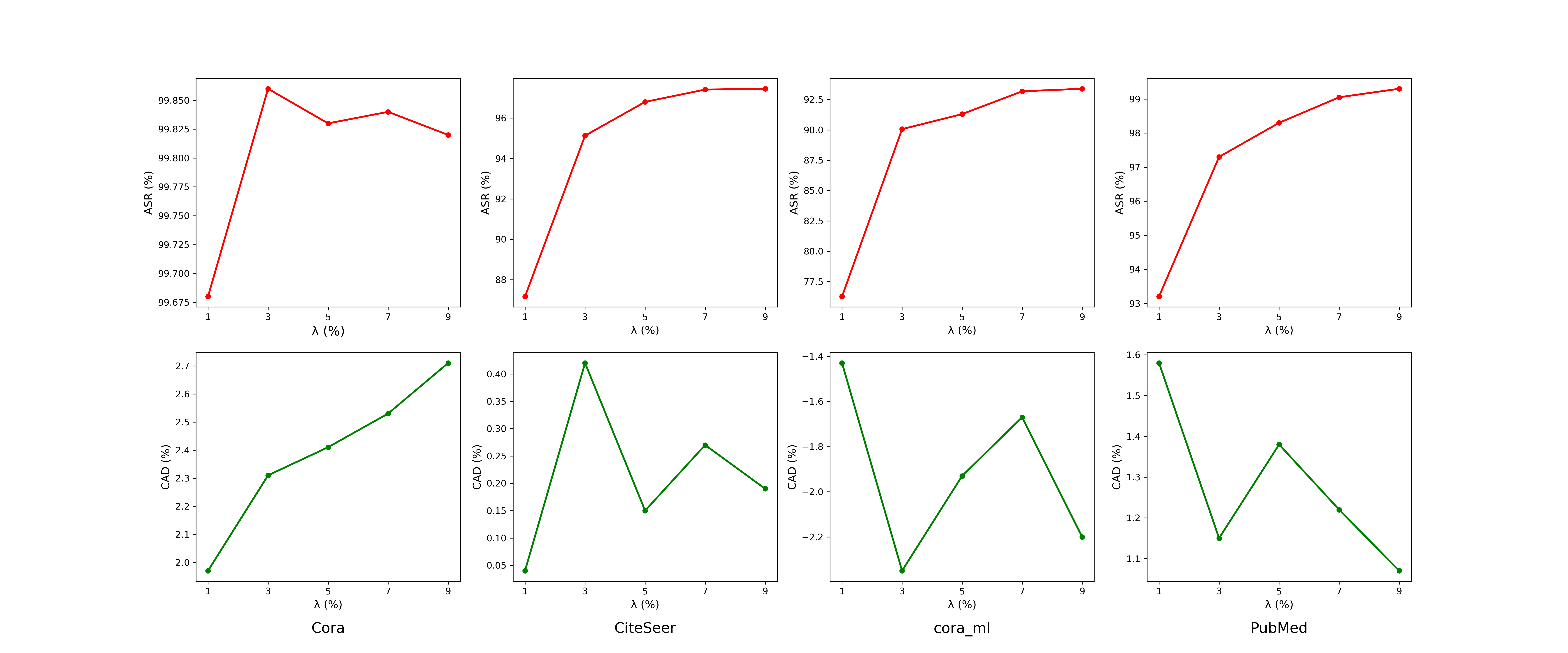}
\caption{\label{fig:size}The impact of ifferent feature combination sizes on ASR and CAD.}
\end{figure}

\subsubsection{Impact of trigger size}

In this section, we investigate the effect of different feature combination sizes $k = \lambda d$ on ASR and CAD, where d denotes the number of node feature dimensions. As shown in Fig. \ref{fig:size}. The horizontal axis of each plot represents the size of $\lambda$, which is taken as 1\%, 3\%, 5\%, 7\%, and 9\%, respectively, and the vertical coordinates of the top four plots represent the ASR, while the vertical coordinates of the bottom four plots represent the CAD. Similarly to the poisoning rate, the ASR increases accordingly with the increase in the size of the feature combinations, and the same trend is observed in all four datasets. CAD, on the other hand, does not show the same trend; on the Cora dataset, as $\lambda$ increases CAD increases accordingly, while on the other three datasets there is not some significant trend. In addition to that, we can find that CBAG only needs to choose a small number of features combination as the trigger pattern to achieve good attack performance. And the CAD is small, which indicates that the prediction accuracy of the backdoor GCN model for benign samples is close to that of the clean model.

\section{Conclusions}
\label{section:6}

In this paper, we propose a black-box backdoor attack against graph convolutional networks, called CBAG, which shows that GCNs are vulnerable to attacks.CBAG considers a more realistic and stealthy attack scenario, which merely modifies the labels of training samples to accomplish backdoor injection. And CBAG uses the node's own features to activate the backdoor, which further enhances the stealthiness. We have conducted extensive experiments on four benchmark datasets, and the results show that CBAG can achieve a high attack success rate even when the poisoning rate does not exceed 5\%. Therefore, in future work, we will focus on further exploring backdoor vulnerabilities in GCNs and exploring effective backdoor defense methods for GCNs.

\bibliographystyle{unsrt}
\bibliography{sample}

\end{document}